\documentclass[letterpaper, 10 pt, conference]{ieeeconf}  

\IEEEoverridecommandlockouts                              

\pdfoutput=1

\usepackage{amsmath,amsfonts,bm}

\def\eqref#1{equation~\ref{#1}}

\def\1{\bm{1}}

\DeclareMathAlphabet{\mathsfit}{\encodingdefault}{\sfdefault}{m}{sl}
\SetMathAlphabet{\mathsfit}{bold}{\encodingdefault}{\sfdefault}{bx}{n}

\def\gD{{\mathcal{D}}}

\def\gH{{\mathcal{H}}}

\def\gK{{\mathcal{K}}}

\def\gM{{\mathcal{M}}}

\def\gP{{\mathcal{P}}}

\newcommand{\R}{\mathbb{R}}

\newcommand{\Null}{\mathbf{Null}}
\newcommand{\True}{\mathbf{True}}

\let\labelindent\relax  
\usepackage{color,xcolor}
\usepackage{epsfig}
\usepackage{graphicx}

\usepackage[export]{adjustbox}
\usepackage{array}
\usepackage{booktabs}
\usepackage{colortbl}
\usepackage{wrapfig}
\usepackage{hhline}
\usepackage{multirow}
\usepackage{subcaption} 
\usepackage{wrapfig}
\usepackage{floatflt}
\usepackage{siunitx}

\usepackage{amsmath,amsfonts,amssymb,amsthm}
\usepackage{mathtools}  
\usepackage{bm}
\usepackage{nicefrac}
\usepackage{microtype}
\usepackage[T1]{fontenc}
\usepackage[ansinew]{inputenc}

\usepackage{changepage}
\usepackage{extramarks}
\usepackage{fancyhdr}
\usepackage{lastpage}
\usepackage{setspace}
\usepackage{soul}
\usepackage{xspace}

\usepackage{url}

\usepackage{enumerate}
\usepackage{todonotes} 
\usepackage{enumitem}  

\usepackage{titlesec}

\usepackage{makecell}

\usepackage{algorithm}
\usepackage[noend]{algpseudocode}

\usepackage{framed}
\definecolor{shadecolor}{gray}{0.95}

\usepackage{xparse}
\usepackage{etoolbox}

\DeclareMathAlphabet{\mathbbb}{U}{bbold}{m}{n}

\newcolumntype{L}[1]{>{\raggedright\let\newline\\\arraybackslash\hspace{0pt}}m{#1}}
\newcolumntype{C}[1]{>{\centering\let\newline\\\arraybackslash\hspace{0pt}}m{#1}}
\newcolumntype{R}[1]{>{\raggedleft\let\newline\\\arraybackslash\hspace{0pt}}m{#1}}

\usepackage{pifont}
\newcommand{\eqn}[1]{Equation~\ref{#1}}
\newcommand{\fig}[1]{Figure~\ref{#1}}
\newcommand{\tbl}[1]{Table~\ref{#1}}

\newcommand{\ignore}[1]{}

\makeatletter
\DeclareRobustCommand\onedot{\futurelet\@let@token\@onedot}
\def\@onedot{\ifx\@let@token.\else.\null\fi\xspace}

\def\ie{i.e\onedot}

\makeatother

\definecolor{MyDarkBlue}{rgb}{0,0.08,1}
\definecolor{URL}{HTML}{0000EE}
\definecolor{MyDarkGreen}{rgb}{0.02,0.6,0.02}
\definecolor{MyDarkRed}{rgb}{0.8,0.02,0.02}
\definecolor{MyDarkOrange}{rgb}{0.40,0.2,0.02}
\definecolor{MyPurple}{RGB}{111,0,255}
\definecolor{MyRed}{rgb}{1.0,0.0,0.0}
\definecolor{MyGold}{rgb}{0.75,0.6,0.12}
\definecolor{MyDarkgray}{rgb}{0.66, 0.66, 0.66}
\definecolor{JiayuanColor}{rgb}{0.60,0.43,0.48}

\newcommand{\model}{KALM\xspace}

\newcommand{\modelfull}{\textit{Keypoint Abstraction using Large Models for Object-Relative Imitation Learning}\xspace}
\newcommand{\myparagraph}[1]{\noindent\textbf{#1}}

\theoremstyle{plain}

\newcommand{\draweropen}{\textsc{DrawerOpen}\xspace}
\newcommand{\drawerclose}{\textsc{DrawerClose}\xspace}
\newcommand{\buttonside}{\textsc{ButtonSide}\xspace}
\newcommand{\buttontop}{\textsc{ButtonTop}\xspace}
\newcommand{\leverpull}{\textsc{LeverPull}\xspace}

\setlist[itemize,1]{leftmargin=\dimexpr 26pt-.2in}
\setlist[enumerate,1]{leftmargin=\dimexpr 26pt-.2in}
\usepackage{dblfloatfix}
\usepackage{cuted} 
\usepackage{capt-of}

\makeatletter
\let\NAT@parse\undefined
\makeatother
\usepackage{hyperref}  

\renewcommand{\paragraph}[1]{\noindent {\bf #1}}

\title{Keypoint Abstraction using Large Models \\for Object-Relative Imitation Learning}

\author{Xiaolin Fang$^\text{*1}$, Bo-Ruei Huang$^\text{*12}$, Jiayuan Mao$^\text{*1}$, Jasmine Shone$^\text{1}$, \\
Joshua B. Tenenbaum$^\text{1}$, Tom\'as Lozano-P\'erez$^\text{1}$, Leslie Pack Kaelbling$^\text{1}$\\%
$^\text{1}$Massachusetts Institute of Technology $^\text{2}$National Taiwan University\\
{\tt\small \{xiaolinf,boruei,jiayuanm,jasshone,jbt,tlp,lpk\}@csail.mit.edu}
}

\begin{document}

\maketitle
\vspace{-1em}
\footnotetext{*: indicates equal contribution.}

\begin{abstract}
Generalization to novel object configurations and instances across diverse tasks and environments is a critical challenge in robotics. Keypoint-based representations have been proven effective as a succinct representation for capturing essential object features, and for establishing a reference frame in action prediction, enabling data-efficient learning of robot skills. However, their manual design nature and reliance on additional human labels limit their scalability. In this paper, we propose \model, a framework that leverages large pre-trained vision-language models (LMs) to automatically generate task-relevant and cross-instance consistent keypoints. \model distills robust and consistent keypoints across views and objects by generating proposals using LMs and verifies them against a small set of robot demonstration data. Based on the generated keypoints, we can train keypoint-conditioned policy models that predict actions in keypoint-centric frames, enabling robots to generalize effectively across varying object poses, camera views, and object instances with similar functional shapes. Our method demonstrates strong performance in the real world, adapting to different tasks and environments from only a handful of demonstrations while requiring no additional labels. Website: \color{URL}{\url{https://kalm-il.github.io/}}.
\end{abstract}

\section{Introduction}

A long-standing goal in robotics is to develop learning mechanisms that allow efficient acquisition of robot skills across a wide range of tasks, requiring a feasible amount of data while generalizing effectively to different object poses and even instances. Such generalization is crucial for enabling robots to perform robustly in diverse environments. One common strategy for improving data efficiency is to leverage abstractions. Researchers have explored object-centric~\cite{zhu2023viola}, part-centric~\cite{liu2024composable}, and keypoint-centric representations~\cite{gao2021kpam2} for tasks spanning rigid-body manipulation to deformable object handling, such as ropes and clothes. Among these, keypoints offer a versatile abstraction for many robotic tasks, capturing essential object features with a low-dimensional encoding.

\begin{figure}[t]
    \centering
    \includegraphics[width=\linewidth]{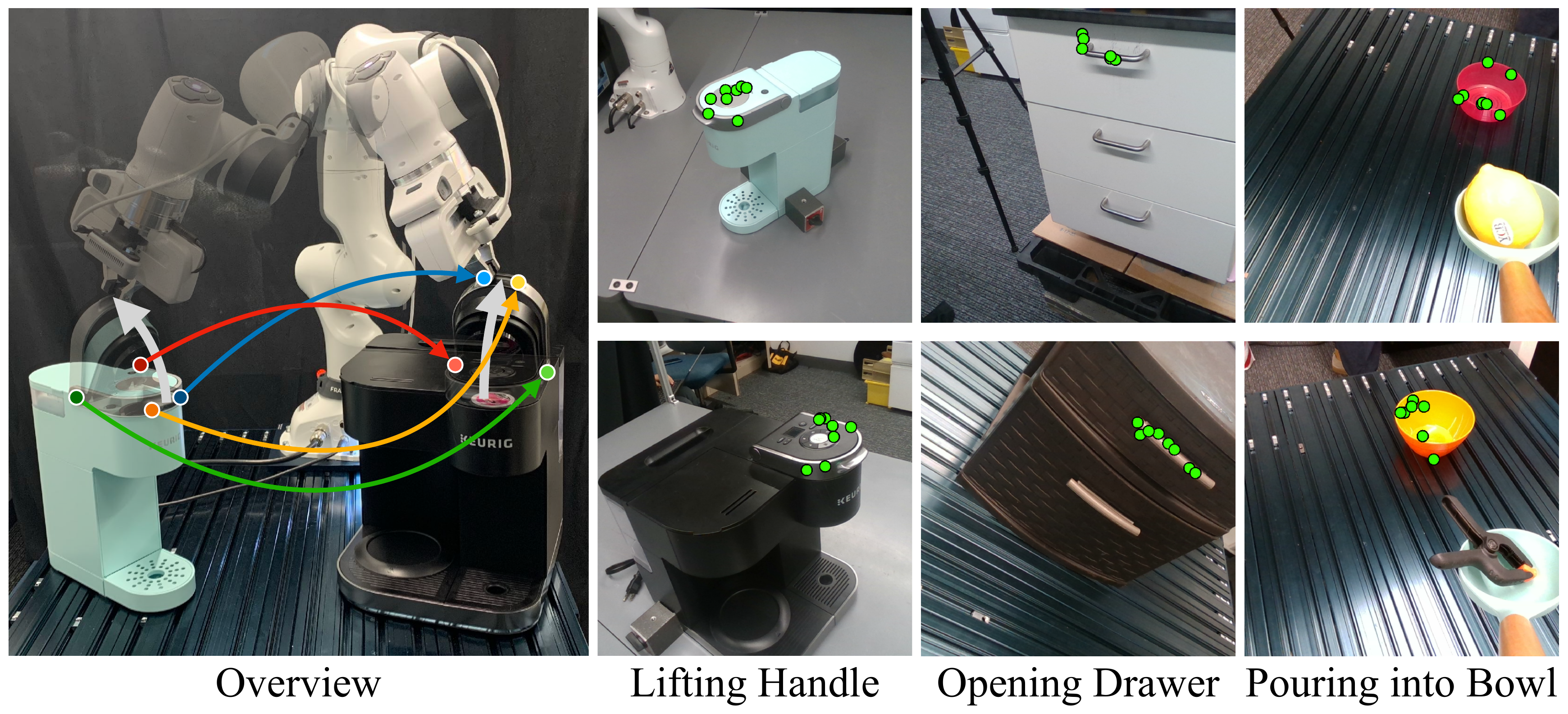}
    \caption{\myparagraph{\modelfull (\model).}
    \model is a framework that distills keypoint abstraction by prompting and verifying keypoint proposals from large pre-trained models using a small amount of robot demonstration data, which is used to train a keypoint-conditioned policy model. Our method demonstrates strong generalization on multiple real-world manipulation tasks with only 10 demonstrations and no additional labeling effort.}
    \label{fig:teaser}
\end{figure}
However, despite their potential for data efficiency and generalization, constructing a good keypoint representation can be tedious, both during training and at test time. Such training typically requires human experts to design task-specific keypoints, while deploying keypoint-centric models in real-world necessitates visual modules capable of detecting these keypoints, which often requires additional data collection and human annotations.

\definecolor{coarse}{HTML}{F27200}
\definecolor{fine}{HTML}{017100}
\definecolor{verify}{HTML}{663366}
\definecolor{phi}{HTML}{B51700}

\begin{figure*}[t]
    \centering
    \includegraphics[width=\linewidth]{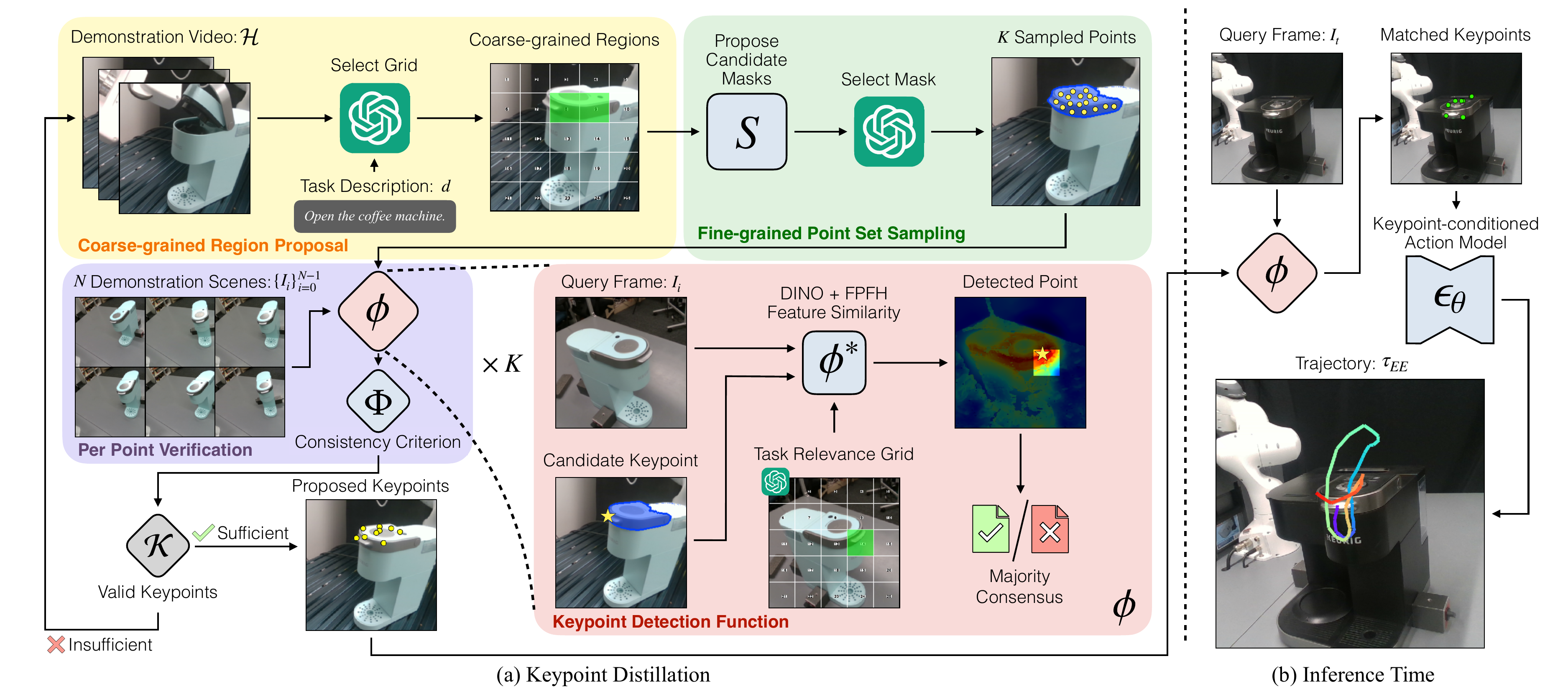}
    \vspace{-1.5em}
    \caption{\myparagraph{\model overview.} 
    \myparagraph{(a) Keypoint distillation.}
    Given a demonstration video and a task description, we prompt a VLM to generate a \textcolor{coarse}{coarse-grained region proposal}, which is refined into a \textcolor{fine}{fine-grained point set} via image segmentation models and VLMs.
    We use a \textcolor{phi}{keypoint detection function $\phi$} to identify keypoint correspondences across a handful of demonstration trajectories.
    The final keypoints set is selected based on \textcolor{verify}{correspondence consistency} verification. These keypoints are used for training a keypoint-conditioned action model.
    \myparagraph{(b) Inference time.}
    Given a new scene, the \textcolor{phi}{keypoint detection function $\phi$} localizes the distilled keypoints. The learned keypoint-conditioned action prediction model generates an object-relative end-effector trajectory based on the keypoint positions and features.
    }
    \label{fig:pipeline}
\end{figure*}

    %

On the other hand, recent advances in large pre-trained models offer a promising alternative for the automatic construction of keypoint-centric representations, but their reliability remains a concern.
Previous attempts to use these models for unsupervised keypoint extraction \cite{huang2024rekep} have exclusively relied on the proposals of task-relevant objects or keypoints from pre-trained models and are therefore error-prone.
Furthermore, it is unclear how these models, usually trained on little or no robotics data, can be aligned with the constraints and uncertainty in the physical environment.
Our key insight is that an appropriate keypoint abstraction should satisfy two essential criteria: task relevance and cross-instance consistency. Task relevance ensures that the keypoints directly support the specific robotic task at hand. Cross-instance consistency ensures that these keypoints are robustly identifiable across multiple instances: from different camera views, over objects having similar functional shapes, in different poses. We hope to use a small amount of robot demonstration data (5 to 10 demonstrations) to align the large pre-trained models to the robotics domain.

In this paper, we propose \modelfull (\model), a framework that distills keypoints by prompting large pre-trained models and verifying the proposed keypoints based on a small number of robot demonstration trajectories. For each new task, given a single seeding demonstration video and a short natural language task description, we prompt a vision-language model to identify candidate object parts that are task-relevant. These candidates are further refined through querying image segmentation models into a candidate set of keypoints. The final keypoint set is selected based on cross-instance consistency computed across available demonstration trajectories of the task.
Once the keypoints are identified, we train a diffusion policy model conditioned on these keypoints and their features to generate robot trajectories relative to the object keypoints. Our system demonstrates strong generalization across real-world settings, adapting to changes in environment and object instances. 

Overall, our contributions are:  First, we propose to distill task-relevant and cross-instance consistent keypoints from large pre-trained models with a combined proposal and verification process. Second, with the extracted keypoints, we build a keypoint-centric, object-relative action representation based on keypoint features and their derived local frames that can be learned by diffusion policy models from a few demonstrations. Our keypoint-conditioned policy model allows robots to generalize their learned behaviors across different environments and object configurations, improving task performance under diverse conditions.
\section{Method}
Illustrated in \fig{fig:pipeline}, our proposed method distills keypoints through an iterative procedure, alternating between candidate proposals using large pre-trained models and verification based on a small robot demonstration dataset. 

Once the keypoints are distilled, we use their visual and geometric features to train a keypoint-conditioned diffusion model for generating object-relative robot actions.

At inference time, our method detects the corresponding points of the previously distilled keypoints in the observed image, predicts the robot actions relative to the detected points, and finally transforms it back into the world frame for execution.

\subsection{Problem Formulation}
\label{ssec:hypothesis}
\vspace{-0.15em}

Formally, for each skill $\alpha$, we require a single video of the robot successfully executing the task $\gH^\alpha$, a handful of demonstration trajectories (5 to 10) $\gD^\alpha$, and a natural language task description $d^\alpha$. Each trajectory $D_i^\alpha$ contains an initial observation of the scene, represented as a calibrated RGBD image $I^\alpha_i$ and a robot joint trajectory $\tau^\alpha_i$. No additional labels such as keypoints are required.

Our first goal is to distill a set of keypoints that are task-relevant and consistent across observed images in all demonstration trajectories $\{ I^\alpha_i \}$. We represent this keypoint set using 3D locations and their features in the first frame of the demonstration video, denoted as $\gK^\alpha$. The second goal is to learn a trajectory prediction model, based on the distilled keypoints and demonstration trajectories $\gD^\alpha$. In this work, we assume the trajectory of each skill is segmented into two distinct phases: an approaching phase, during which the robot moves freely towards the object, and an execution phase. The trajectory prediction model only predicts the execution trajectory, focusing only on the actions needed to manipulate the objects. The segmentation of the trajectory can be done through thresholding the distance to the closest keypoint, or through methods checking more fine-grained trajectory statistics~\cite{ke20243ddiffuseractorpolicy, gervet2023act3d3dfeaturefield, shridhar2022peract}. For brevity, in the following, we omit the superscript $\alpha$ when there are no ambiguities. 

\subsection{Keypoint Desiderata}
\label{ssec:criteria}
\vspace{-0.15em}

In our framework, the keypoints serve both as an abstraction of the observational input and the basis of an object-relative action frame. To ensure that the abstraction is effective and the frame is robust to changes in the environment, we define two criteria: task relevance and cross-instance consistency. 

\paragraph{Task relevance.} 
To allow generalization to different scene configurations of the same task, the keypoints must be task-relevant. For example, for the task of lifting the handle of a coffee machine, the points on the handle are ideal candidates whereas those on the water reservoir are not because the latter varies across different machines and does not directly support the task completion.

Given a demonstration $D_i = \langle I_i, \tau_i \rangle$, the skill description $d$, as well as a keypoint $k$ and the corresponding position $p_{k,i} = \phi(k, I_i)$ in $I_i$, a pretrained vision-language model will implicitly assign a score $\psi$ to this keypoint $k$, denoted as $\psi(k, D_i, d, p_{k,i})$. Our overall goal is to find a set of keypoints $\gK$ such that $\psi(k, D_i, d, p_{k,i})$ is high for all training demonstrations $D_i$.

\paragraph{Cross-instance consistency.} 
Furthermore, it is essential that the keypoints are consistently identifiable across observations regardless of the object pose, camera view, or detailed shape of objects. For example, within a task-relevant object part, a point on the corner may be favored over one on a plain surface, due to its saliency and a lower degree of ambiguity.

We evaluate the task relevance of a keypoint by leveraging pre-trained vision-language models, and check the cross-instance consistency using a keypoint detection function $\phi$. Our goal is to find a set of keypoints that are both task-relevant and consistently identifiable.

\subsection{Keypoint Proposal and Verification}
\label{ssec:proposal}

Our keypoint proposal and verification pipeline works in three steps. First, we prompt a pre-trained vision language model (VLM) to select task-relevant image regions. Within the regions, we generate queries to image segmentation models to generate candidate object part masks, which are further ranked and selected by a second query to the VLM. We sample fine-grained keypoint proposals within the selected mask and score them based on consistency across all query images from the training demonstration set. This process will either return a set of final keypoints or declare failure, leading to another iteration of keypoint proposal. The overall process is illustrated in Figure~\ref{fig:pipeline}a and Algorithm~\ref{alg:proposal-and-verification}.

\paragraph{Coarse-grained region proposal $\textit{VLM}_1(\gH, d)$.}
Our input to the VLM consists of a sequence of images $\gH$ showing a single demonstration video of the robot executing the task, along with a natural language description of the skill $d$ (e.g., ``open the top drawer''). 
We aim to identify the regions of interest in the initial image, $I_0^\gH$, associated with this video. 
We present $I_0^\gH$ with an overlaid grid, where each grid cell is indexed by a unique text label, and query the VLM to select the grid indices corresponding to the task-relevant regions.

In addition to the grid index, we employ zero-shot chain-of-thought~\cite{wei2022chain} prompting, encouraging the VLM to generate a textual description of the target object, and its highlighted part before generating the final prediction. 

\begin{algorithm}[tp]
\caption{Keypoint Proposal and Verification Pipeline}
\label{alg:proposal-and-verification}
\begin{algorithmic}[1]
\Require Skill description $d$, training set $\gD = \{ \langle I_i, \tau_i \rangle \}_{i=0}^{N-1}$, demonstration video $\gH = \{I_t^\gH\}_{t=0}^{|\gH|-1}$
\Ensure{Set of proposed keypoints $\gK$ and their matched point in the training set $\gP :=\phi(k, I_i), \forall k \in \gK, i \in I_i$}
\While {$\True$}
\State $\mathcal{K} \leftarrow \emptyset$; $\mathcal{P} \leftarrow \emptyset$
\State $\mathcal{K}_\textit{VLM} \leftarrow \textit{VLM}_1(H, d)$ \Comment{Region proposal}
\State $\gM_s \leftarrow \cup_{k \in \gK_\textit{VLM}} S(I_0^\gH, k)$ \Comment{Image segmentation}
\State $m \leftarrow \textit{VLM}_2(I_0^\gH, k, \gM_\textit{S})$ \Comment{Mask selection}
\State $\gK_{\textit{FPS}} \leftarrow \textit{FPS}(I_0^\gH, m)$ \Comment{Fine-grained point sampling}
\For {each $k' \in \gK_{\textit{FPS}}$}
\If{$\frac{1}{N} \sum_{i=0}^{N-1} \mathbbb{1}\left[ \mathbb{\phi}(k', I_i) \neq \Null \right] \geq \1 - \delta$}
\State $\gK = \gK \cup \{k'\}$ \Comment{Add $k'$ to the output set}
\State $\gP = \gP \cup \{\phi(k', I_i) \mid \forall I_i \}$
\EndIf
\EndFor
\If {$\frac{|\gK|}{|\gK_\textit{FPS}|} \geq \gamma$}
\State $\mathbf{break}$
\EndIf
\EndWhile
\end{algorithmic}
\end{algorithm}

\paragraph{Mask proposal $\textit{S}(I_0^\gH, k)$ and $\textit{VLM}_2(I_0^\gH, k, \gM_{\textit{S}})$.} 
Next, for each candidate coarse-grained region (represented as a grid cell in $I_0^\gH$), we generate a uniformly distributed set of query points within the cell, and use a point-prompted image segmentation model~\cite{kirillov2023segment} to generate a set of object-part masks $\mathcal{M}_{\textit{S}}$ for each of these query points. We apply standard non-maximum suppression to filter out masks with significant overlaps and discard those with low confidence scores.

Subsequently, we provide the conversation history from the previous VLM query, along with the image $I_0^\gH$ overlaid with all the detected masks $\mathcal{M}_{\textit{S}}$, as input of a second VLM query $\textit{VLM}_2$. 
The VLM is tasked with selecting a single mask from $\mathcal{M}_{\textit{S}}$ that contains the potential task-relevant keypoints. The output of this step is a mask $m$ on the image $I_0^\gH$. 
By leveraging the VLM's understanding of both the task context and the object-part segmentation masks, we obtain a more detailed representation of the task-relevant region.

\paragraph{Fine-grained Point Sampling $\textit{FPS}(I_0^\gH, m)$.}  
Given the input RGBD image $I_0^\gH$ and the selected mask $m$ from the second VLM query, we 
apply Farthest Point Sampling~\cite{eldar1997farthest} to the 3D points located inside the mask $m$ to generate $N_c$ candidate points $\gK_\textit{FPS}$. We found Farthest Point Sampling to be effective at generating diverse spatially distributed candidate keypoints that are geometrically salient, such as those located on part boundaries. 
These points are often visually distinct and thus tend to be consistently identifiable. 
We will then select a set of keypoints from this set through our cross-instance consistency verification.

\paragraph{Keypoint detection function $\phi(k, I)$.} 
Each candidate keypoint $k$ in $\gK_\textit{FPS}$ is internally represented by its position $p^k_0$ and feature vector $F_\textit{ref}(p^k_0)$ in the initial scene $I_0^\gH$. We need to detect their corresponding points in the training set observations $\{I_i\}$. We treat the keypoint detection task as a correspondence matching problem, where the goal is to identify the matching keypoint in a new input image $I$. The keypoint detection function can be written as $\phi(k, I) \rightarrow p$, where $I \in \R^{H\times W \times 4}$ is a query RGBD image, and $p \in \R^3$ is the 3D position of the keypoint in the query scene. When no matched keypoint is found in $I$, $\phi$ returns $\Null$. 

This is accomplished using a scoring function $\phi^*$ based on the per-point feature representation in $I$. Formally, given the keypoint position $p^k_0$ in $I_0^\gH$, the function $\phi$ returns the position $p$ in $I$ that maximizes the following score: 
\begin{equation}
\phi^*(p, p^k_0) = \langle F_\textit{q}(p), F_\textit{ref}(p^k_0) \rangle,
\label{eq:similarity}
\end{equation}
where $\langle \cdot, \cdot \rangle$ denotes the cosine similarity between two vectors.
We also use the same VLM pipeline to obtain a coarse image grid of the target keypoint on the new input image $I$ and discount points outside the mask in this matching process.

To enhance robustness, we implement a local group-based matching strategy that randomly samples $N' = 8$ neighboring points within a radius of $r = 0.02 \, \text{m}$ around $p_0$, computes the corresponding match for each of them in the input image $I$, and declares a non-match if there is no majority consensus among the sampled points (\ie, if the majority of the matched points in $I$ are not within close proximity of each other).

\paragraph{Cross-instance consistency verification.}
The previous steps generate a set of proposed keypoints $\gK_\textit{FPS}$ from the VLM and their corresponding points (or non-match) in each image of the demonstration trajectories. We evaluate each keypoint based on $\Phi$ that requires a successful correspondence matching in a majority of demonstration trajectories:
\vspace{-2.25em}
\begin{center}
\begin{equation}
    \Phi(k) := ~\frac{1}{N} \sum_{i=0}^{N-1} \mathbbb{1}\left[ \mathbb{\phi}(k, I_i) \neq \Null \right] \geq 1 - \delta,
\end{equation}
\end{center}
\vspace{-0.5em}
where the acceptance factor $\delta = 0.3$ in our experiment.

Our success criteria for the proposal set $\gK_\textit{FPS}$ is that the majority of points in this set should be consistent candidates: ${\sum_k \Phi(k)}/{|\gK_\textit{FPS}|} \geq \gamma$. If there are sufficient consistent candidates in the proposal set $\gK_\textit{FPS}$, our algorithm returns those consistent candidates as our final selected keypoints $\gK$. Otherwise, we will discard all proposal keypoints and re-prompt VLM to generate another mask and set of proposal points. This step is important for verifying that the VLM selected part is both task-relevant and consistently identifiable.

\subsection{Learning Keypoint-Centric Trajectory Prediction Models}
\label{ssec:diffuser}

From the keypoint proposal and verification process, we have determined a set of keypoints $\gK$, which captures the most salient and task-relevant object parts for the skill $\alpha$. 
For each demonstration trajectory $D_i$, we also have a set of corresponding points $\gP_i$ from the previous keypoint detection step. Conditioned on the sparse keypoint locations and features of these detected points, we directly generate a trajectory $\tau_\textit{EE}$ for the 6 DoF pose of the robot's end-effector using the Diffuser~\cite{janner2022diffuser}, a trajectory-level diffusion model. Internally, the Diffuser learns a score function $\epsilon_\theta(\tau_\textit{EE} \mid \mathbf{C})$ parameterized by $\theta$, which captures the gradient of the data distribution over $\tau_\textit{EE}$, where $\mathbf{C}$ is the conditional input to the diffuser.

We have two key design choices here to facilitate generalization: using a sparse keypoint-based input, and having actions predicted in a keypoint-centric, object-relative coordinate frame. Specifically, the model only takes the keypoint locations and their features as input, which leverages the keypoint abstraction to obtain invariance to task-irrelevant distractions, such as background and view changes. Meanwhile, we predict the actions relative to the center of these keypoints. The object-relative nature of the design makes our model invariant to changes in the absolute pose of the camera and the objects.



\subsection{Implementation}
\label{ssec:implementeation}
We use GPT-4o~\cite{openai2024gpt4o} as our VLM and Segment-Anything Model~\cite{kirillov2023segment} (SAM) as our image segmentation model.
In the similarity function $\phi^*$, we use a combination of pre-trained image features (DINO~\cite{oquab2023dinov2} with FeatUp~\cite{fu2024featup}) and analytic 3D features Fast Point Feature Histograms (FPFH~\cite{rusu2009fast}). The overall (cosine) similarity in \eqn{eq:similarity} is defined as:
\begin{equation}
\begin{aligned}
 \langle F_\textit{q}(p), F_\textit{ref}(p^k_0)\rangle & = \lambda_1 \cdot \langle F^{\textit{DINO}}_\textit{q}(p), F^{\textit{DINO}}_\textit{ref}(p^k_0) \rangle \\
& + \lambda_2 \cdot \langle F^{\textit{FPFH}}_\textit{q}(p), 
F^{\textit{FPFH}}_\textit{ref}(p^k_0) \rangle,
\end{aligned}
\end{equation}
where $\lambda_1=0.75$ and $\lambda_2=0.25$ in our experiments. 

For the trajectory prediction model, we employ Diffuser~\cite{janner2022diffuser}. The trajectory $\tau_\textit{EE}$ is represented as a sequence of $H = 48$ poses. For each pose, we use a 10-dimensional vector that includes a three-dimensional location, 6-dimensional rotation vector~\cite{zhou2019rotation6d}, and one dimension for the gripper. The input keypoint feature to the Diffuser is DINO and FPFH as in the keypoint detection function. We optimize the model using standard diffusion loss functions. The diffusion model estimates the conditional distribution $p(\tau_\textit{EE} \mid \mathbf{C})$, where $\mathbf{C} = \{\langle p_k, F^{\textit{DINO}}_k, F^{\textit{FPFH}}_k \rangle \}_{k \in \mathcal{K}}$ represents the set of matched keypoints in $D_i$, $\tau_\textit{EE}$ is the end-effector trajectory, which is used for inference time denoising.

\subsection{Inference-time Pipeline}
\label{ssec:inference}

At inference time, given a new scene image $I$, we begin by running the keypoint detector $\phi(k, I)$ for all keypoints $k \in \gK$, extracting their corresponding position and feature vectors. We then use the learned Diffuser to generate a set of end-effector trajectories. Starting from randomly initialized trajectories sampled from Gaussian noise, the model employs the backward process which iteratively denoises the noisy trajectories through gradient steps guided by the score function $\epsilon_\theta$ under given conditions.

Note that the learned trajectory starts relatively close to the target object. We need to ensure reachability and collision-free motion in the environment. Similar to previous work that uses diffusion models as trajectory samplers~\cite{fang2023dimsam}, we use a motion planner (bi-directional RRT~\cite{lavalle2001randomized}) to check whether there is a feasible path to the initial pose of the task trajectory $\tau_{\textit{EE}}^{t=0}$. If the motion planner returns no valid path, we will test the next predicted trajectory, until all generated trajectories in the set are exhausted, when the algorithm returns failure. Otherwise, we move the end-effector to $\tau_{\textit{EE}}^{t=0}$ using the approaching path returned by the motion planner, and start $\tau_{\textit{EE}}$ execution.



\section{Experiment}

In this section, we want to study the following questions:
\begin{itemize}
    \item Is the sparse keypoint abstraction sufficient for the conditional diffusion model to predict valid trajectories with only a limited amount of demonstration data?
    \item Does the keypoint abstraction improve data efficiency compared to the baselines?
    \item Does the iterative proposal and verification procedure distill appropriate keypoints for the action model learning?
\end{itemize}

We study the data efficiency of keypoint abstraction in a simulation environment due to the difficulty in collecting a large amount of data for training all baselines in the real world. We also compare different keypoint proposal methods in the real world by measuring the success rate and generalization of the keypoint-conditioned policies.

\subsection{Simulation Experiments in Meta-World}

\begin{table*}[t]
\sisetup{mode=text,table-alignment-mode=format,table-number-alignment=center,uncertainty-mode = separate,table-format = 2.2(2.2)}
\centering\small
\begin{adjustbox}{max width=\textwidth}
\begin{tabular}{@{}clccccc@{}}
\toprule
&& \multicolumn{2}{c}{Without Keypoints}& \multicolumn{3}{c}{With Keypoints}\\ \cmidrule(lr){3-4} \cmidrule(lr){5-7}
\multicolumn{1}{c}{\multirow{2}{*}{\#Demos}}&\multicolumn{1}{c}{\multirow{2}{*}{Tasks}}& RGB & RGBD& RGB& \multicolumn{2}{c}{RGBD} \\ \cmidrule(lr){3-3} \cmidrule(lr){4-4} \cmidrule(lr){5-5} \cmidrule(lr){6-7}
&& Diffuser~\cite{janner2022diffuser} & 3D Diffuser Actor~\cite{ke20243ddiffuseractorpolicy}& Diffuser& 3D Diffuser Actor& \textbf{\model (Ours)} \\ \midrule
\multirow{5}{*}{10}
& \draweropen & \tablenum{30.00 \pm 8.29} & \tablenum{30.00 \pm 6.53} & \tablenum{62.00 \pm 1.41} & \tablenum{29.33 \pm 4.64} & \textbf{\tablenum[detect-weight]{77.00 \pm 2.94}} \\ 
& \drawerclose & \tablenum{50.00 \pm 1.41} & \tablenum{53.67 \pm 4.19} & \tablenum{83.33 \pm 2.62} & \tablenum{50.67 \pm 6.60} & \textbf{\tablenum[detect-weight]{92.33 \pm 0.47}} \\ 
& \buttonside & \tablenum{32.67 \pm 2.49} & \tablenum{37.67 \pm 2.05} & \tablenum{49.67 \pm 11.09} & \tablenum{38.67 \pm 1.25} & \textbf{\tablenum[detect-weight]{79.67 \pm 1.25}} \\ 
& \buttontop & \tablenum{19.00 \pm 3.56} & \tablenum{21.00 \pm 4.90} & \tablenum{28.00 \pm 8.04} & \tablenum{21.00 \pm 4.32} & \textbf{\tablenum[detect-weight]{97.33 \pm 0.47}} \\ 
& \leverpull & \tablenum{10.67 \pm 3.30} & \tablenum{8.67 \pm 2.36} & \tablenum{21.33 \pm 1.70} & \tablenum{10.33 \pm 4.19} & \textbf{\tablenum[detect-weight]{61.67 \pm 6.13}} \\ 
\bottomrule
\end{tabular}
\end{adjustbox}
\caption{\textbf{Few-shot learning in Meta-World.} We evaluate our method, \model, on five manipulation tasks in the Meta-World~\cite{yu2019meta} simulator, using Diffuser~\cite{janner2022diffuser} and 3D Diffuser Actor~\cite{ke20243ddiffuseractorpolicy} as baselines, along with ablation studies on keypoint usage. Our method consistently outperforms these baselines, demonstrating that keypoints serve as an effective abstraction.}
\label{tab:simulator_result}
\end{table*}

\definecolor{KeypointPink}{RGB}{255, 0, 255}

\begin{figure}[t]
    \centering
    \captionsetup{size=scriptsize}
    \begin{subfigure}[b]{0.24\linewidth}
        \centering
        \includegraphics[width=\textwidth]{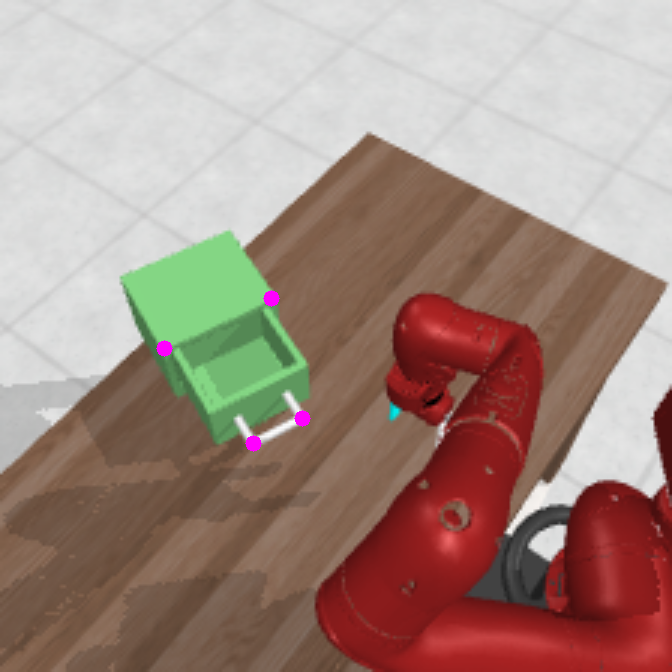}
        \captionsetup{justification=raggedright,format=hang}
        \caption{\draweropen\\\drawerclose}
        \label{fig:sim_vis_drawer}
    \end{subfigure}
    \begin{subfigure}[b]{0.24\linewidth}
        \centering
        \includegraphics[width=\textwidth]{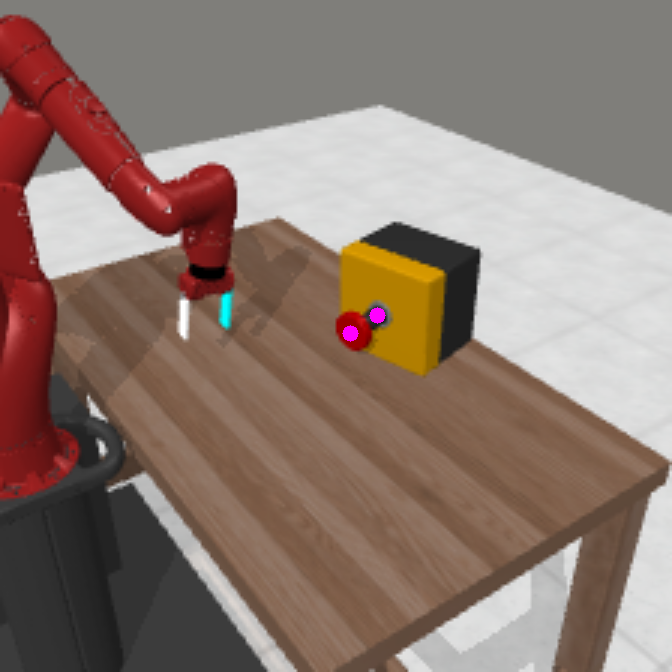}
        \caption{\buttonside\\\hspace{\textwidth}}
        \label{fig:sim_vis_button}
    \end{subfigure} 
    \begin{subfigure}[b]{0.24\linewidth}
        \centering
        \includegraphics[width=\textwidth]{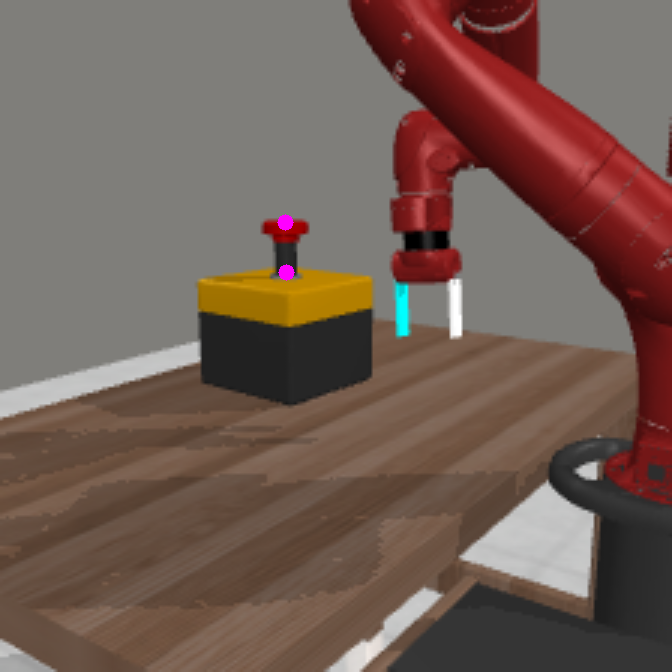}
        \caption{\buttontop\\\hspace{\textwidth}}
        \label{fig:sim_vis_button_topdown}
    \end{subfigure}
    \begin{subfigure}[b]{0.24\linewidth}
        \centering
        \includegraphics[width=\textwidth]{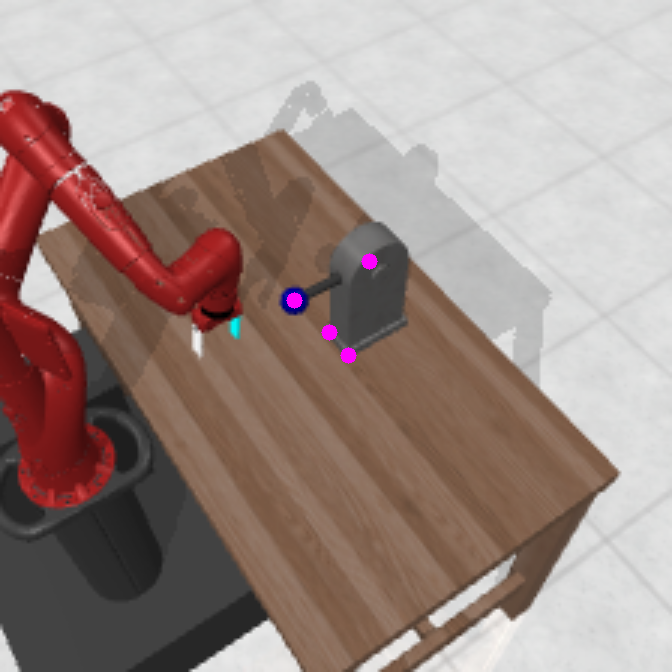}
        \caption{\leverpull\\\hspace{\textwidth}}
        \label{fig:sim_vis_lever}
    \end{subfigure}
    \captionsetup{size=small}
    \caption{\myparagraph{Testing tasks in Meta-World~\cite{yu2019meta} simulator.} 
    We evaluate on 5 tasks in Meta-World with randomized camera and object poses, necessitating the generalization of policies across observational changes.
    Keypoints are marked in \textcolor{KeypointPink}{pink} for visualization.}
    \label{fig:sim_vis}
\end{figure}

In this section, we compare our method with other baselines with different input spaces and network architectures using the Meta-World~\cite{yu2019meta} simulator, focusing on the efficacy of different representations in terms of their data efficiency.

\myparagraph{Setup.} We evaluate on 5 tasks: \draweropen, \drawerclose, \buttonside, \buttontop, and \leverpull, as shown in \fig{fig:sim_vis}. For each task, we provide an oracle set of keypoints by manually labeling the XML files and computing their 2D or 3D locations in the scene using known object and camera poses. We train and test on scenes with varying camera and object poses. The object poses are uniformly sampled on the table with 2D translations and rotations. The camera viewpoint is sampled around the object with randomized angle and distance with elevation in $[0, \frac{\pi}{2}]$, azimuth in $[-\frac{\pi}{2}, \frac{\pi}{2}]$, and distance in $[1.5, 2]$. We exclude random camera angles where the arm obstructs the object in the initial scene.


\myparagraph{Baselines.}
We compare our method against 4 baselines.
\begin{itemize}
\item \textbf{Diffuser~\cite{janner2022diffuser} (RGB)} generates a 6 DoF end-effector trajectory based on an initial RGB observation of the scene and the camera extrinsic, using a 1D convolutional network. We use DINO~\cite{oquab2023dinov2} visual encoder and finetune it at training time to handle the discrepancy between the simulator and real-world images on which DINO is trained.
\item \textbf{3D Diffuser Actor~\cite{ke20243ddiffuseractorpolicy} (RGBD)} builds a 3D representation from pre-trained visual features (CLIP~\cite{radford2021learning}) and the point cloud. It generates actions with a diffusion model conditioned on the 3D representation.
\item \textbf{Diffuser with keypoints}: We provide Diffuser with the 2D position of the keypoints as an additional input.
\item \textbf{3D Diffuser Actor with keypoints}: Similarly, we provide the 3D position of the keypoints to the 3D Diffuser Actor.\end{itemize}

\myparagraph{Few-shot learning results.} 
We present the task success rates of all baselines across 5 tasks trained with only 10 demonstrations in \tbl{tab:simulator_result}. The success rate is averaged over 100 test episodes and the variances are computed across three random seeds.
Overall, the Diffuser (RGB) model struggles to predict accurate 3D trajectories. Nonetheless, we observe a significant performance improvement with the keypoints added. The 3D Diffuser Actor, which employs a transformer backbone, requires more training data. It has poor performance in low-data regimes. By contrast, our keypoint-conditioned diffusion model, taking only sparse inputs of 3D keypoint positions and their visual features, can learn efficiently from very few demonstrations while achieving strong performance. 

\myparagraph{Data efficiency study.} 
\begin{figure}[t]
    \centering
    \includegraphics[width=\linewidth]{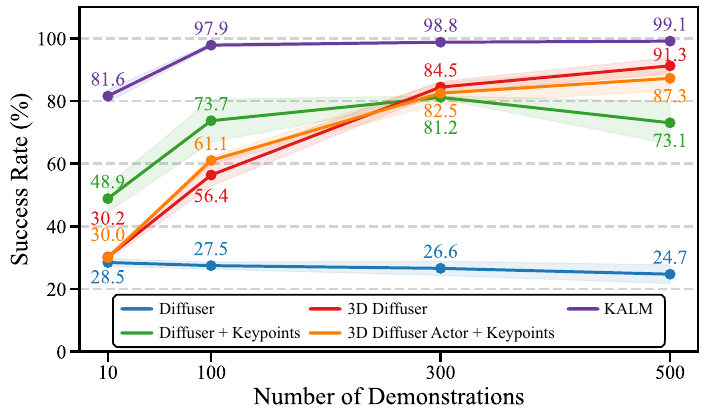}
    \vspace{-1.5em}
    \caption{\myparagraph{Data efficiency.} We measure the average success rate across all 5 tasks, with the number of demonstrations increasing from 10 to 500. 
    Our method, \model, demonstrates superior data efficiency compared to all baselines. 
    }
    \label{fig:data_efficiency}
\end{figure}
We further evaluate the data efficiency by varying the number of demonstrations from 10 to 500. We report the average success rate on 5 tasks in \fig{fig:data_efficiency} (each with three seeds and 100 test episodes). Our method \model achieves superior data efficiency, reaching peak performance using only 100 demonstrations, whereas the 3D Diffuser Actor requires 500 demonstrations to achieve competitive performance. This validates the effectiveness of sparse keypoint abstraction for trajectory prediction.

\subsection{Real-World Experiment on Franka Arm}

In this section, we investigate whether our proposed iterative procedure generates better keypoints which eventually lead to a higher task success rate for a range of different tasks in the real world. We also explore the generalization capability along different dimensions endowed by using keypoint abstraction such as object poses and object instances.

\begin{figure}[t]
    \centering
    \begin{subfigure}[b]{0.31\linewidth}
        \centering
        \includegraphics[width=\textwidth]{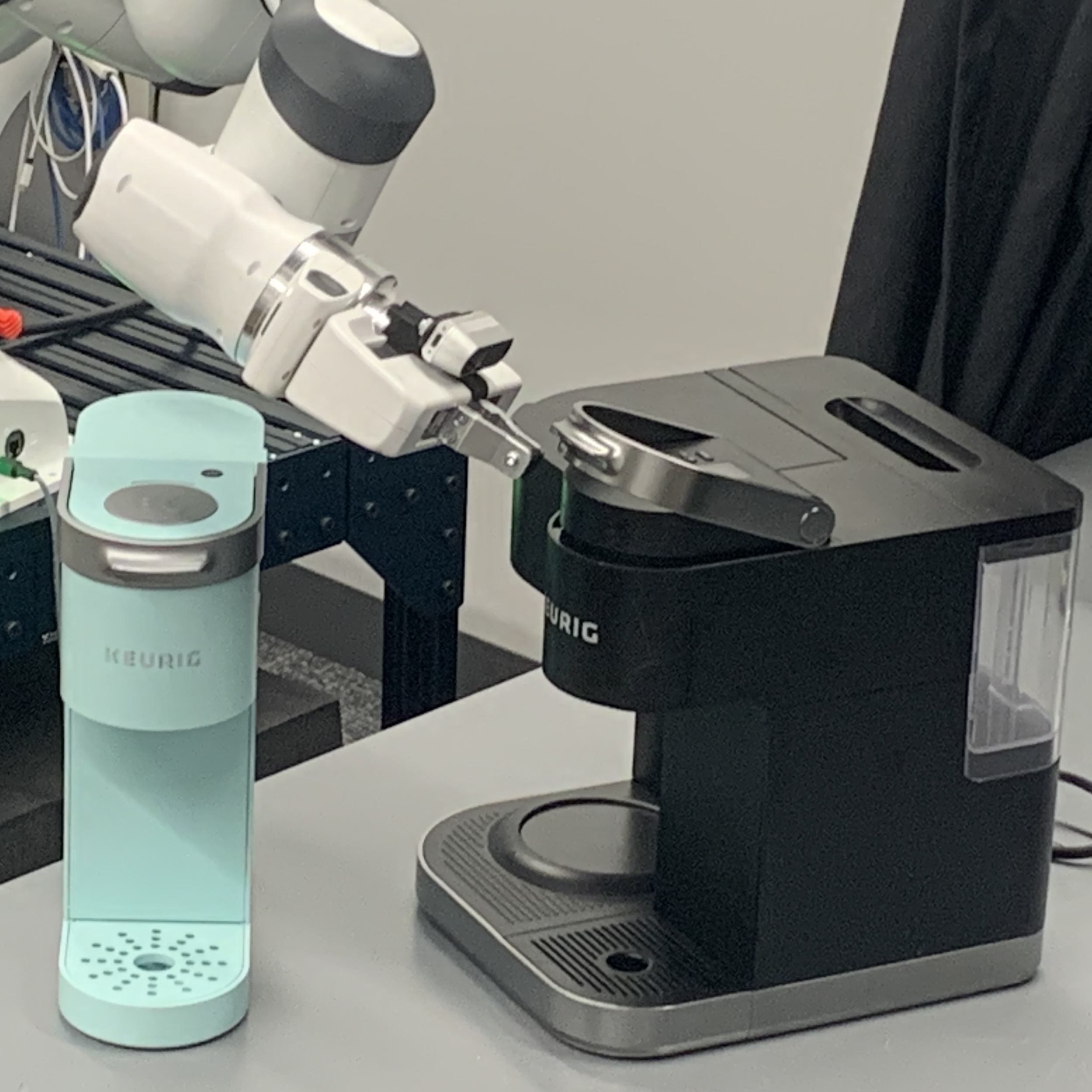}
        \caption{Lifting the handle}
        \label{fig:real_vis_coffee}
    \end{subfigure}
    \begin{subfigure}[b]{0.31\linewidth}
        \centering
        \includegraphics[width=\textwidth]{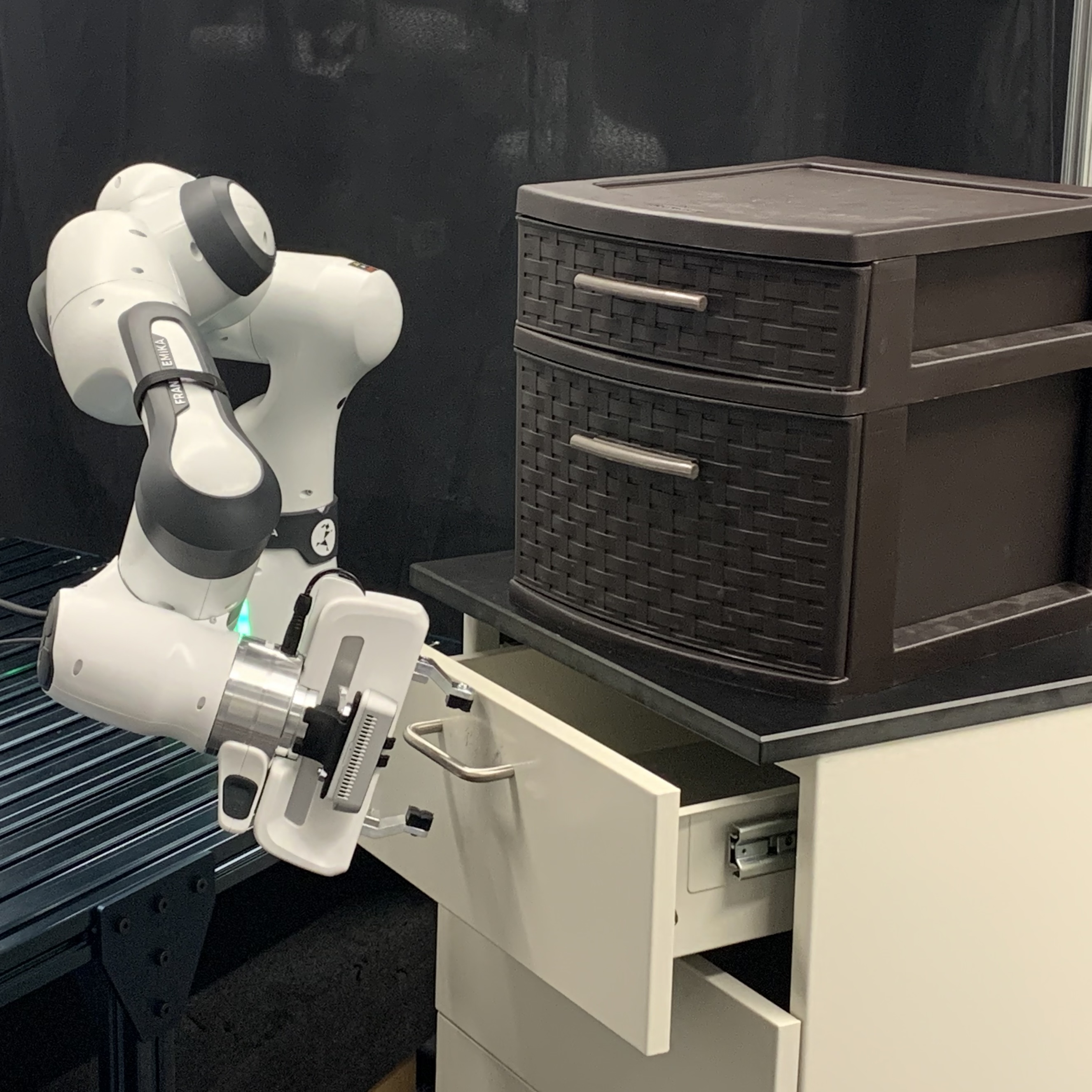}
        \caption{Opening the drawer}
        \label{fig:real_vis_drawer}
    \end{subfigure} 
    \begin{subfigure}[b]{0.31\linewidth}
        \centering
        \includegraphics[width=\textwidth]{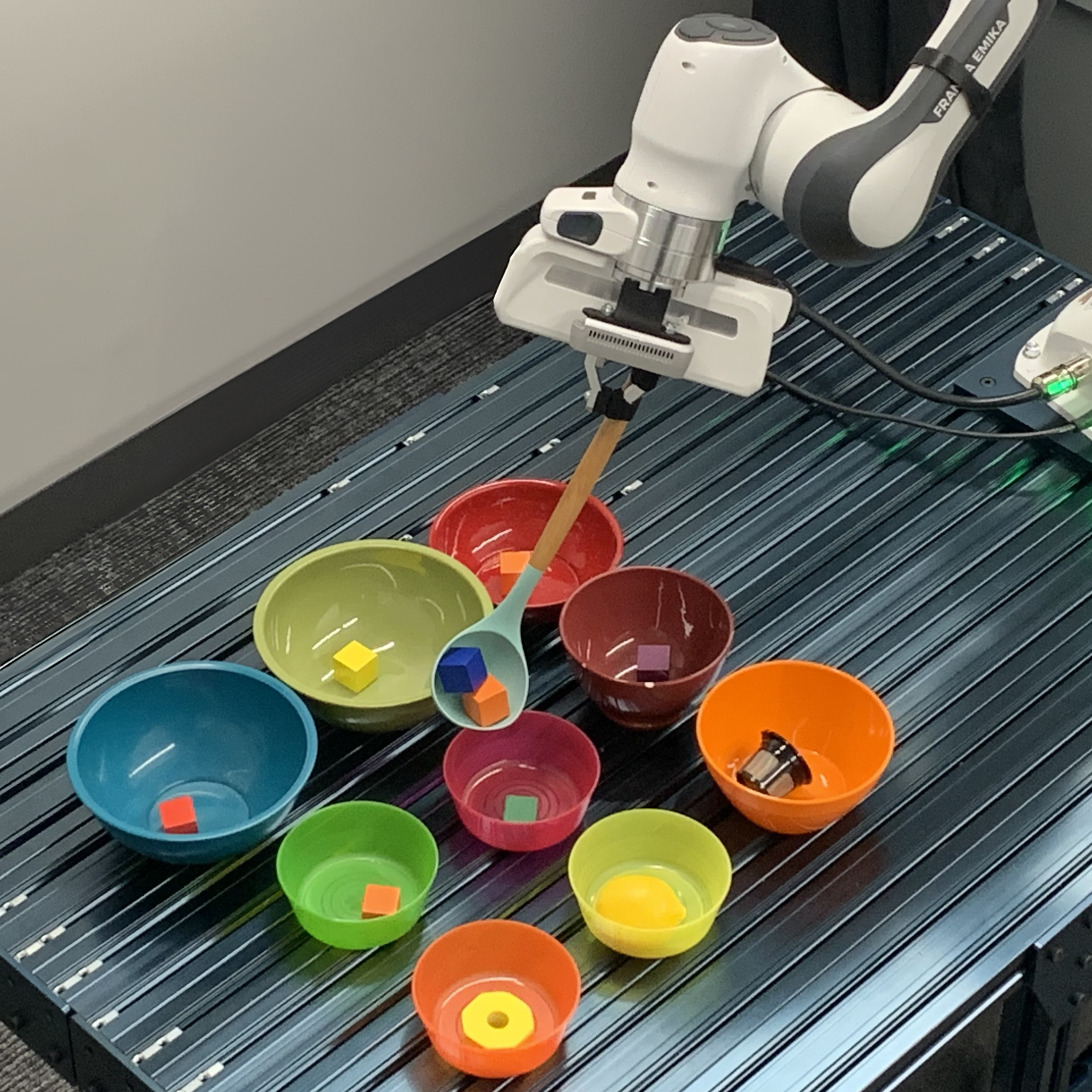}
        \caption{Pouring into a bowl}
        \label{fig:real_vis_pour}
    \end{subfigure}
    \captionsetup{size=small}
    \caption{\myparagraph{Testing tasks in the real world.} 
    We evaluate different methods on three tasks in the real world with different objects at different poses, and with different camera angles. The testing assets are illustrated in the figure.}
    \label{fig:realworld-vis}
\end{figure}

\myparagraph{Setup.}
We carry out real-world experiments on three tasks: 1) Lifting the handle of a coffee machine, 2) Opening the top drawer, and 3) Pouring something into a bowl, as illustrated in \fig{fig:realworld-vis}. 
We use a Franka Research 3 robot arm with a RealSense D435i RGBD camera mounted on the gripper. For each task, we collect 10 demonstrations and capture the initial image using the gripper-mounted camera.

To evaluate the generalization along different dimensions, we vary the environment in terms of camera and object poses (View), and object instances (Cross obj.). 
Note that the objects are never seen during the training of the diffusion model, where changes in camera and object poses are also applied.

\definecolor{KeypointPink}{RGB}{255, 0, 255}

\begin{table}[t]
\centering \small
\setlength{\tabcolsep}{6.1pt}
    \begin{tabular}{lcccc}
    \toprule
    &\multicolumn{2}{c}{w/o. Verification} & \multicolumn{2}{c}{\model (Ours)} \\
    \cmidrule(lr){2-3} \cmidrule(lr){4-5}
    
    Tasks & View & Cross Obj. & View & Cross Obj.    \\
    \midrule
    Lifting Handle & 1/10 & 0/10 & 9/10 & 6/10 \\
    Opening Drawer & 2/10 & 0/10 & 6/10 & 7/10 \\
    Pouring into Bowl & 6/10 & 2/10 & 8/10 & 6/10 \\
    \bottomrule
    \end{tabular}
    \caption{\myparagraph{Real-world experiments success rate.}  We evaluate all models in varying camera and object poses (View), as well as on unseen objects (Cross obj.). Our method demonstrates strong performance under view changes and showcases generalization to objects that are not seen during diffusion training time.}
    \label{tab:real_overall_result}
\end{table}

\myparagraph{Baseline.} To validate the efficacy of the verification procedure, we compare our method against a baseline that directly prompts VLMs to select a set of keypoints without performing cross-instance consistency verification, referred to as ``w/o. Verification''. Specifically, we make another query to VLM after obtaining the fine-grained point sampled $\gK_\textit{FPS}$, and ask it to select $N_k=8$ keypoints, as the final output. 

\myparagraph{Few-shot learning results.}
We carry out 10 repeated runs for each task. 
For a fair comparison, we run the baseline and \model with the same set of 10 setups: the robot is reset to the same initial pose before executing the predicted trajectory, and we reset the objects as close as possible if they were moved by the robot. \tbl{tab:real_overall_result} shows the overall results.

Overall, our method consistently outperforms the baseline, highlighting the importance of cross-instance consistency verification. We observe that the baseline struggles with localizing corresponding keypoints in testing scenes. 
For example, in the handle-lifting task, the predicted trajectory is usually centered on the wrong parts of the coffee machine due to keypoint localization failures.
Even when the keypoint detection is accurate, the predicted trajectories are significantly worse. We attribute the reason to the poor detection during keypoint distillation at training time, which leads to a lower-quality dataset $\gP$ for training the keypoint-conditioned action prediction model, resulting in reduced performance.

With the object-relative action representation, our method achieves strong performance under view changes, abstracting away the absolute position of the camera or object. This enables our method to perform strongly with only 10 demonstrations and no additional labels. We believe this is an efficient way to scale up generalizable skill learning.

\section{Related Work}
\label{sec:related_work}

\paragraph{Abstractions for action representations}
Abstractions over raw sensorimotor interfaces have been shown to enhance data-efficient learning and facilitate generalization. Various forms of abstractions have been studied, including contact points~\cite{trinkle1991framework, ji2001planning, lee2015hierarchical, cheng2022contact}, objects~\cite{diuk2008object, devin2018objectcentricrl, veerapaneni2020entity, wang2021handeye, yuan2022sornet, mao2022pdsketch, zhu2023viola}, object parts~\cite{aleotti-icar2011, vahrenkamp2016part, myers2015affordance, liu2023composable}, keypoints~\cite{Manuelli2019KPAM, gao2021kpam2, qin2020keto, GIFT}, and other shape representations~\cite{yodo, simeonov2022neural,shen2023F3RM}. These spatial abstractions can serve as direct inputs to data-driven models~\cite{wang2021handeye, gao2021kpam2, zhu2023viola}, or be used to create coordinate frames~\cite{niekum2012learning, liu2023composable}. Typically, these models represent short-horizon robot behaviors and can be sequentially composed~\cite{qin2020keto, wang2024grounding} in a fixed order, or integrated into high-level planning algorithms~\cite{garrett2021integrated}. Our work leverages keypoint-based representations for few-shot imitation learning, and focuses on acquiring such representations automatically without additional data and labels.

\paragraph{Finding keypoint correspondences.} Finding functional correspondences between objects~\cite{lai2021functional} has been extensively explored in robotic manipulation, using both analytical methods~\cite{curvature_primal_sketch, sascha2014generalizing, diego2018transfer, interation_warping} and data-driven methods~\cite{simeonov2022neural, thompson2021shape, Manuelli2019KPAM, gao2021kpam2, GIFT, yodo}. However, the keypoints derived from these approaches are typically tailored to specific tasks, requiring human annotations of task-relevant keypoints at training time. Furthermore, although data-driven methods offer better generalization to novel objects, they require additional data for training, such as labeled keypoints or additional object datasets. In contrast, our method eliminates the need for human-labeled keypoints by leveraging off-the-shelf large pre-trained models for vision-language grounding and visual recognition, enabling automatic discovery of task-relevant and cross-instance consistent keypoints.

\myparagraph{Vision-language models for robotics.}
A large body of research has focused on utilizing pre-trained language and vision-language models for robotic manipulation, by generating action plans~\cite{saycan2022arxiv, liang2023code}, specifying motion constraints~\cite{huang2023voxposer,huang2024rekep}, and directly predicting robot movements~\cite{hu2023look, fangandliu2024moka}. The primary limitation of these methods is their dependence on primitive sets of low-level controllers or on motions that can be easily described by simple analytical expressions. Additionally, because these approaches rely on one-shot solutions generated by pre-trained models, they are prone to errors produced by these models. Other work proposes generating reward functions~\cite{ma2023eureka} or planning models~\cite{liu2021learning} and can learn new policies from data, but they do not tackle the issue of data efficiency.

\section{Conclusion}
We propose \model, a framework that distills task-relevant keypoints by iteratively prompting LMs and verifying consistency using a small amount of data. We use these keypoints as an abstraction to learn a keypoint-conditioned policy model that predicts object-relative robot actions, leveraging the keypoints as observational abstractions and local action frames. Our simulated and real-world experiment shows that our keypoint representation enables data-efficient learning and facilitates generalization to changes in the scene.

\paragraph{Acknowledgements.} We gratefully acknowledge support from NSF grant 2214177; from AFOSR grant FA9550-22-1-0249; from ONR MURI grant N00014-22-1-2740; and from ARO grant W911NF-23-1-0034; from MIT Quest for Intelligence; from the MIT-IBM Watson AI Lab; from ONR Science of AI; and from Simons Center for the Social Brain. Any opinions, findings, and conclusions or recommendations expressed in this material are those of the authors and do not necessarily reflect the views of our sponsors.

\bibliographystyle{IEEEtran}
\bibliography{references}

\end{document}